\documentclass{article}
\usepackage{ijcai21}

\usepackage{times}

\usepackage{soul}
\usepackage{url}
\usepackage[hidelinks]{hyperref}
\usepackage[utf8]{inputenc}
\usepackage[small]{caption}
\usepackage{graphicx}
\usepackage{amsmath}
\usepackage{booktabs}
\usepackage{epsfig}
\usepackage{amssymb}
\usepackage{wrapfig}
\usepackage[ruled, linesnumbered]{algorithm2e}
\usepackage{xcolor}
\usepackage{bbm}
\usepackage{algorithmic}
\usepackage{subcaption}
\usepackage{microtype}      
\usepackage{makecell}
\usepackage{multirow}
\usepackage{mathtools}
\usepackage{algorithmic}
\usepackage{booktabs} 
\usepackage{diagbox}
\usepackage{threeparttable}
\usepackage{bm}
\title{Semi-Supervised Deep Ensembles for Blind Image Quality Assessment}

\author{
Zhihua Wang$^1$\footnote{Contact Author}\and
Dingquan Li$^2$\and
Kede Ma$^{1}$\\
\affiliations
$^1$Department of Computer Science, City University of Hong Kong\\
$^2$ Peng Cheng Laboratory\\
\emails
zhihua.wang@my.cityu,edu.hk,
lidq01@pcl.ac.cn,
kede.ma@cityu.edu.hk
}

\begin{document}

\maketitle

\begin{abstract}
Ensemble methods are generally regarded to be better than a single model if the base learners are deemed to be ``accurate'' and ``diverse.'' 
Here we investigate a semi-supervised ensemble learning method to produce generalizable blind image quality assessment models. We train a multi-head convolutional network for quality prediction by maximizing the accuracy of the ensemble (as well as the base learners) on labeled data, and the disagreement (\textit{i.e.}, diversity) among them on unlabeled data, both implemented by the fidelity loss. We conduct extensive experiments to demonstrate the advantages of employing unlabeled data for BIQA, especially in model generalization and failure identification.
\end{abstract}

\section{Introduction}


Data-driven blind image quality assessment (BIQA) models~\cite{bosse2017deep,ma2017end} employing deep convolutional networks (ConvNets) have achieved unprecedented performance, as measured by the correlations with human perceptual scores. However, the performance improvement may be doubtful due to the conflict between the small scale of test IQA datasets and the large scale of ConvNet parameters,  heightening the danger of poor generalizability. 
Ensemble learning, which aims to improve model generalizability by making use of multiple ``accurate'' and ``diverse'' base learners~\cite{zhang2012ensemble}, is a promising way of alleviating this conflict, and has great potentials in outputting generalizable BIQA models. 

Researchers in the field of machine learning have contributed many brilliant ideas to  ensemble learning, \textit{e.g.}, bagging~\cite{breiman1996bagging}, boosting~\cite{freund1997decision}, and negative correlation learning~\cite{liu1999ensemble}. 
These methods are mainly demonstrated under the supervised learning setting where the training labels are given. 
In many real-world problems, plenty of unlabeled data are mostly freely accessible, while the collection of human labels is prohibitively labor-expensive. For example, in BIQA, acquiring the  mean opinion score (MOS) of one image involves effort of $15$ to $30$ subjects~\cite{sheikh2006statistical}. 

In this paper, we combine labeled and unlabeled data to train deep ensembles for BIQA in the semi-supervised setting~\cite{chapelle2006semi,chen2018semisupervised}.  
Our method is based on a multi-head ConvNet, where each head corresponds to a base learner and produces a quality estimate. To reduce model complexity, the base learners share considerable amount of early-stage computation and have separate later-stage convolution and fully connected (FC) layers~\cite{lee2015m}. 
The ensemble is end-to-end optimized to trade off two theoretical conflicting objectives - ensemble accuracy (on labeled data) and diversity (on unlabeled data), both implemented through the fidelity loss \cite{tsai2007frank}.
We conduct extensive experiments to show that the learned ensemble performs favorably against a ``top-performing'' BIQA model - UNIQUE \cite{zhang2020uncertainty} in terms of quality prediction on existing IQA datasets, while exhibiting much stronger generalizability in the group maximum differentiation (gMAD) competition \cite{ma2018group}. Moreover, our results further suggest that promoting diversity helps the ensemble spot its corner-case failures,  which is in turn beneficial for subsequent active learning \cite{wang2020active}.


\section{Method}
\label{sec:method}
Let $f^{(i)}$ denotes the $i$-th base learner, which is implemented by a deep ConvNet, consisting of several stages of convolution, batch normalization \cite{ioffe2015batch}, half-wave rectification (\textit{i.e.}, ReLU nonlinearity) and spatial subsampling, followed by FC layers for quality computation. Given an input image $x$, the ensemble method $f$ is simply defined by the average of all base learners:
\begin{align}\label{eq:ensemble}
     f(x) = \frac{1}{M}\sum_{i=1}^{M}f^{(i)}(x), 
\end{align}
where $M$ is the number of base learners used to build the ensemble. A necessary condition for Eq.~\eqref{eq:ensemble} to be valid is that all base learners need to produce quality estimates of the same perceptual scale, which is nontrivial to satisfy when formulating BIQA as a ranking problem \cite{ma2017dipiq}. We will give a detailed treatment of this scale alignment in Sec.~\ref{subsec:se}.

\subsection{Supervised Ensemble Learning for BIQA}
\label{subsec:se}
Following~\cite{ma2017dipiq,zhang2020uncertainty}, we choose the pairwise learning-to-rank (L2R) method for BIQA learning due to its feasibility of training BIQA models on multiple human-rated datasets. Specifically, given a labeled training set $\mathcal{L}$, in which each image $x$ is associated with an MOS $\mu_x$, we convert it into another set $\mathcal{P}_l$, where the input is a pair of images $(x,y)$ and the target output is a binary label
\begin{align}
p =
\begin{cases}
1& \mu_x \ge \mu_y\\
0& \text{otherwise}
\end{cases}.
\end{align}

Under the Thurstone's Case V model \cite{thurstone1927law}, we assume that the true perceptual quality $q(x)$ of image $x$ follows a Gaussian distribution with mean estimated by the ensemble $f(x)$. The probability $p(x, y)$ indicating that $x$ is of higher quality than $y$ can then be computed by
\begin{align}
    \hat{p}(x, y)=\operatorname{Pr}(q(x) \geq q(y))=\Phi\left(\frac{f(x)-f(y)}{\sqrt{2}}\right),
    \label{eq:thurstone}
\end{align}
where $\Phi(\cdot)$ is the standard Normal cumulative distribution function, and the standard deviation (std) is fixed to one. We may alternatively use the $i$-th base learner to estimate the mean of the Gaussian distribution, and compute a corresponding $\hat{p}^{(i)}(x,y)$ by replacing $f$ with $f^{(i)}$ in Eq.~\eqref{eq:thurstone}. We use the fidelity loss \cite{tsai2007frank} to quantify the similarity between two discrete probability distributions:
\begin{align}\label{eq:fidelity}
\ell(p, \hat{p}) = 1 - \sqrt{p\hat{p}} - \sqrt{(1-p)(1-\hat{p})}.
\end{align}
We then define the optimization objective over a mini-batch $\mathcal{B}_l\subset\mathcal{P}_l$ as 
\begin{align}
\label{eq:jointloss}
\begin{split}
\ell_{\mathrm{acc}}(\mathcal{B}_l) = \frac{1}{\left|\mathcal{B}_l\right|}\sum_{x, y \in \mathcal{B}_l}&\Bigg(\ell\left(p(x,y), \hat{p}(x,y)\right)\Bigg. \\
&+\Bigg.\frac{\lambda}{M}\sum_{i=1}^{M} \ell\left(p(x, y), \hat{p}^{(i)}(x,y)\right)\Bigg),
\end{split}
\end{align}
\noindent where the first term is the ensemble loss and the second term is the mean individual loss, respectively.  $\left|\mathcal{B}_l\right|$ denotes the cardinality of $\mathcal{B}_l$. $\lambda$ is set to one by default.

It is noteworthy that the objective in Eq.~\eqref{eq:jointloss} that relies on the Thurstone's model in Eq.~\eqref{eq:thurstone} suffers from the translation ambiguity. As a result, the learned base models $\{f^{(i)}\}_{i=1}^M$ may not live in the same perceptual scale. 
We empirically find three simple tricks that are effective in calibrating the base learners. First, $\ell_2$-normalize the input feature vector to FC layers \cite{wang2017normface,zhang2021continual}, projecting it onto the unit sphere.
Second, remove the biases of the FC layers. Third, batch-normalize the output of the FC layers (\textit{i.e.}, the quality estimate) and share the learnable scale parameter across all base learners (with the bias parameter fixed to zero).

 As an additional note, an alternative way of computing $\hat{p}(x,y)$ by the ensemble is to average the probabilities estimated by the base learners:
 \begin{align*}
    \hat{p}(x, y) =\frac{1}{M} \sum_{i=1}^{M}\hat{p}^{(i)}(x,y).
 \end{align*}

Throughout the paper, we opt for Eq.~\eqref{eq:thurstone} because it is the ensemble prediction (in Eq.~\eqref{eq:ensemble}) that will be used during model deployment, and it gives slightly better quality prediction results.

\subsection{Semi-Supervised Ensemble Learning for BIQA}
We now incorporate unlabeled data for learning BIQA models with the goal of maximizing ensemble diversity. Similar to the supervised setting, we create a second image set $\mathcal{P}_u$ by sampling pairs of images from a large pool of unlabeled images. Inspired by the seminal work of negative correlation learning \cite{buschjager2020generalized,chen2018semisupervised}, we define  diversity as the negative average of the pairwise fidelity losses between all pairs of  base learners: 
\begin{align}
\label{eq:divloss}
 \ell_{\mathrm{div}}(\mathcal{B}_u)= - \frac{1}{\binom{M}{2}\left|\mathcal{B}_u\right|} \sum_{x, y \in \mathcal{B}_u} \sum_{i<j}\ell\left(\hat{p}^{(i)}(x,y), \hat{p}^{(j)}(x,y)\right),
\end{align}
where $\mathcal{B}_u$ is a mini-batch sampled from $\mathcal{P}_u$. Only unordered pairs of base learners are considered due to the symmetry of the fidelity loss (Eq.~\eqref{eq:fidelity}). As there is no standardized definition of diversity, other  implementations may also be plausible, including the  prediction variance:
\begin{align}\label{eq:mse}
\ell_{\mathrm{div}}(x) = -\frac{1}{M}\sum_{i=1}^M\left(f^{(i)}(x) - f(x)\right)^2,
\end{align}
where $f(x)$ is defined in Eq.~\eqref{eq:ensemble} and the negative mean of the fidelity loss between the base learners and the ensemble:
\begin{align*}
 \ell_{\mathrm{div}}(x,y)= - \frac{1}{M}  \sum_{i=1}^M\ell\left(\hat{p}^{(i)}(x,y), \hat{p}(x,y)\right),
\end{align*}
where $\hat{p}(x,y)$ is defined in Eq.~\eqref{eq:thurstone}.

We combine the ensemble accuracy term on labeled data and the ensemble diversity term on unlabeled data to obtain the final objective function:
\begin{align}
   \label{eq:tloss}
   \ell_{\mathrm{semi}}(\mathcal{B}_l, \mathcal{B}_u) = \ell_{\mathrm{acc}}(\mathcal{B}_l) + \gamma \ell_{\mathrm{div}}(\mathcal{B}_u),
\end{align}
where $\gamma$ is the trade-off parameter. In our experiments, it does not hurt to include a diversity term on the labeled $\mathcal{B}_l$ treated as unlabeled~\cite{wang2021longtailed}.

\section{Experiments}
In this section, we first describe the experimental setup, and then present the main results, followed by extensive ablation studies.

\subsection{Experimental Setups}
\label{subsec:experiment_setup}
\noindent\textbf{Implementation Details}. We use the convolutional structure in ResNet-18 \cite{he2016deep} as the backbone, and add one FC layer for multi-head prediction.
The first convolution and the subsequent three residual blocks are shared for each base learner to reduce the model complexity and computational cost~\cite{lee2015m}.
The backbone parameters are initialized with the weights pre-trained on ImageNet \cite{deng2009imagenet}, and the FC parameters are initialized by He's method \cite{he2015delving}. We train the entire method using the Adam optimizer~\cite{kingma2014adam} with a mini-batch size of $16$ for twelve epochs. The initial learning rate is set to $10^{-4}$, which is halved for every epoch. The best parameters are selected according
to the performance on the validation set. 
The images during training are cropped to $384 \times 384$, keeping their aspect ratios, while the test images are fed with the original sizes. 

\begin{figure*}[!t]
	\centering
	\subfloat[Top: $72$, bottom: $48$]{\includegraphics[width=0.23\textwidth]{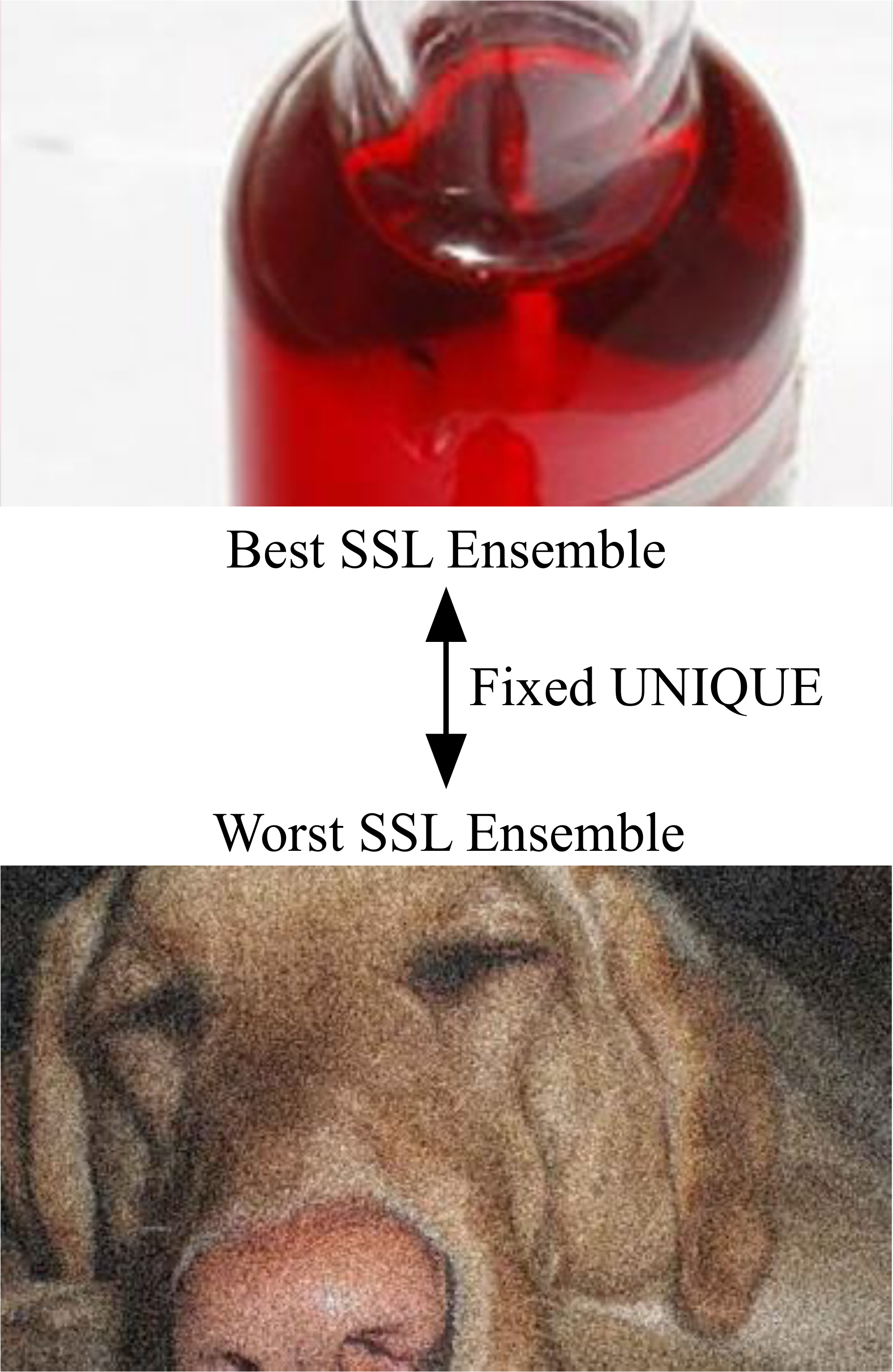}}\hskip.25em
	\subfloat[Top: $72$, bottom: $23$]{\includegraphics[width=0.23\textwidth]{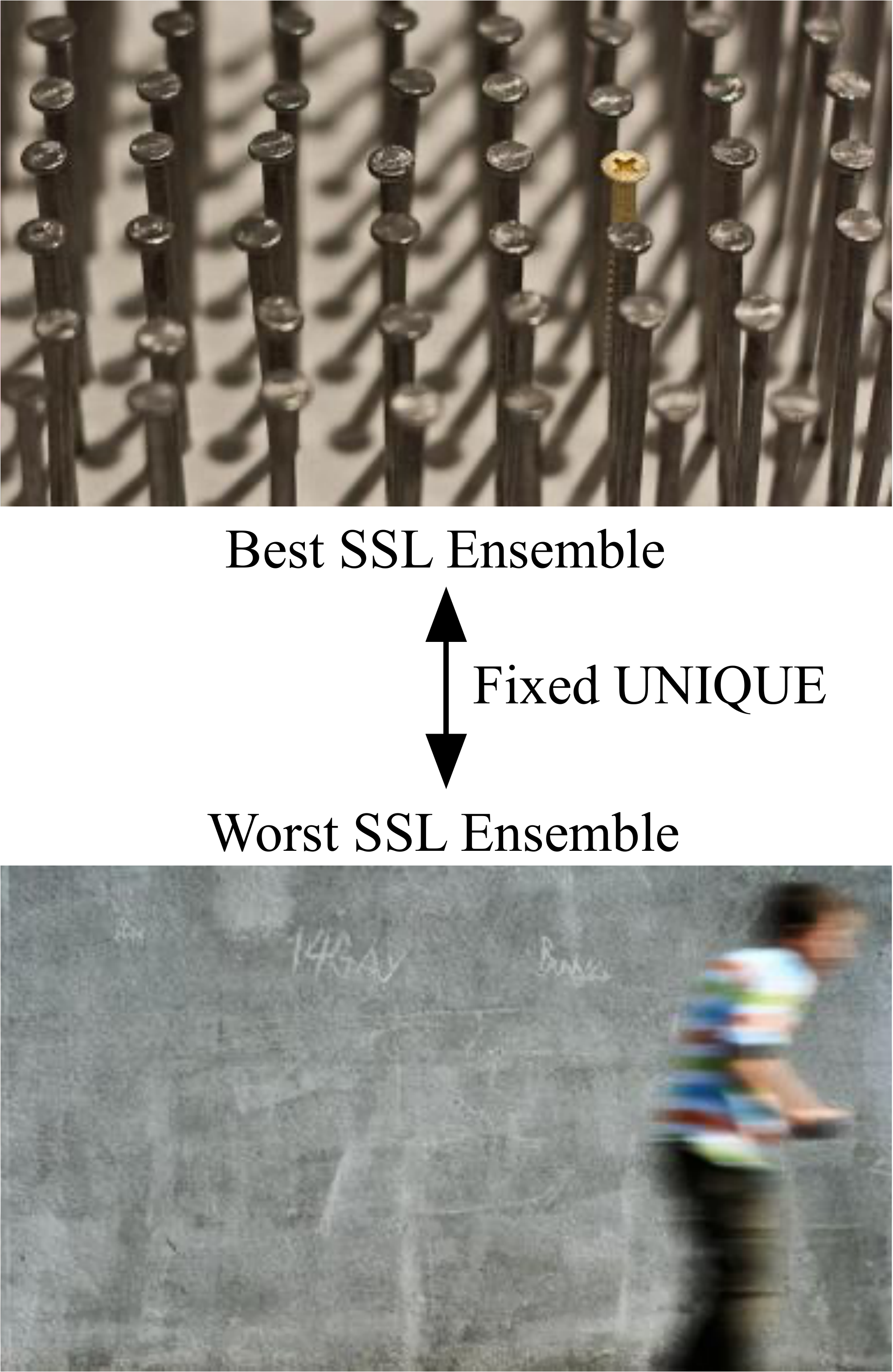}}\hskip.25em
	\subfloat[Top: $62$, bottom: $70$]{\includegraphics[width=0.23\textwidth]{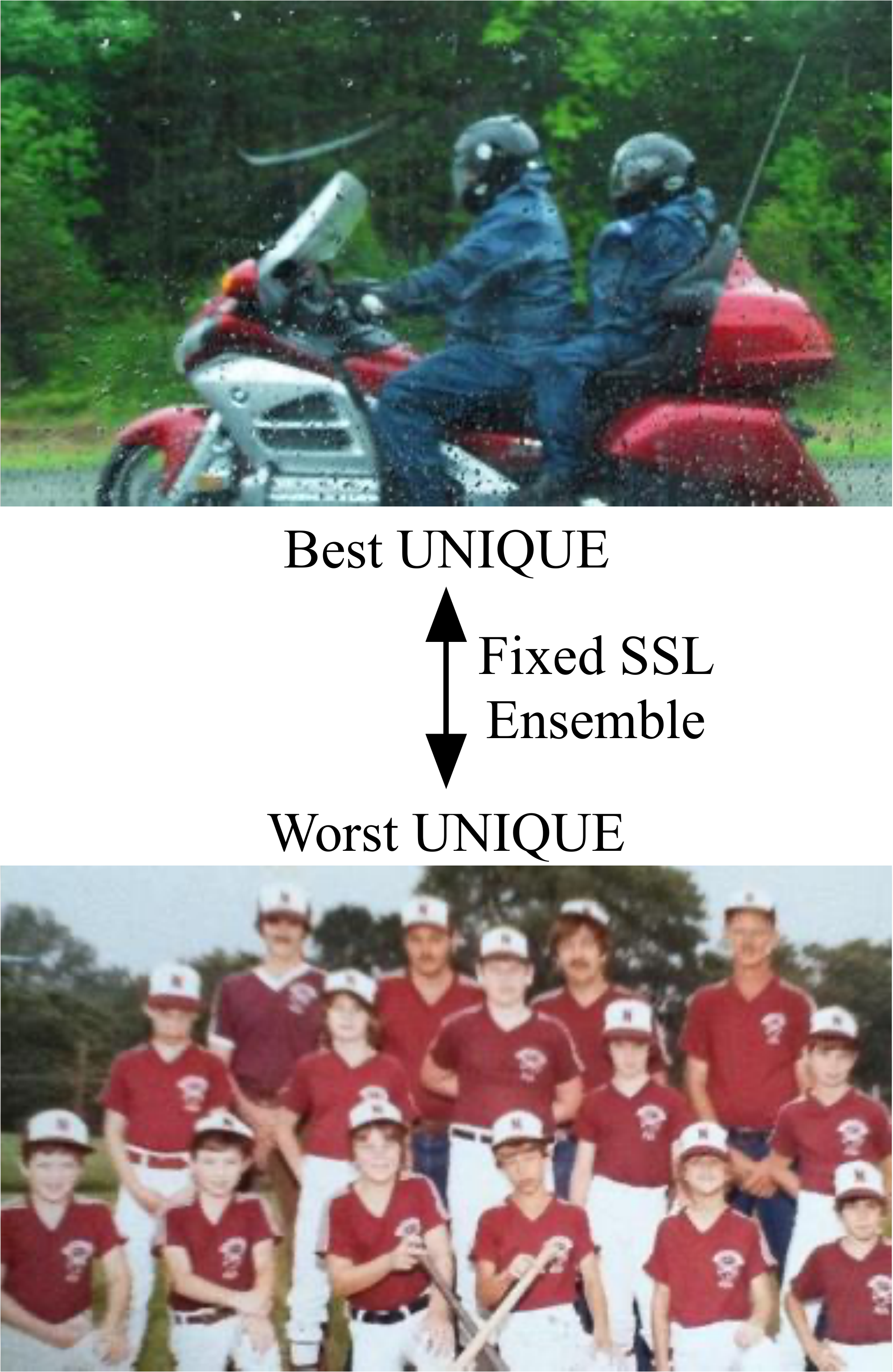}}\hskip.25em
	\subfloat[Top: $72$, bottom: $74$]{\includegraphics[width=0.23\textwidth]{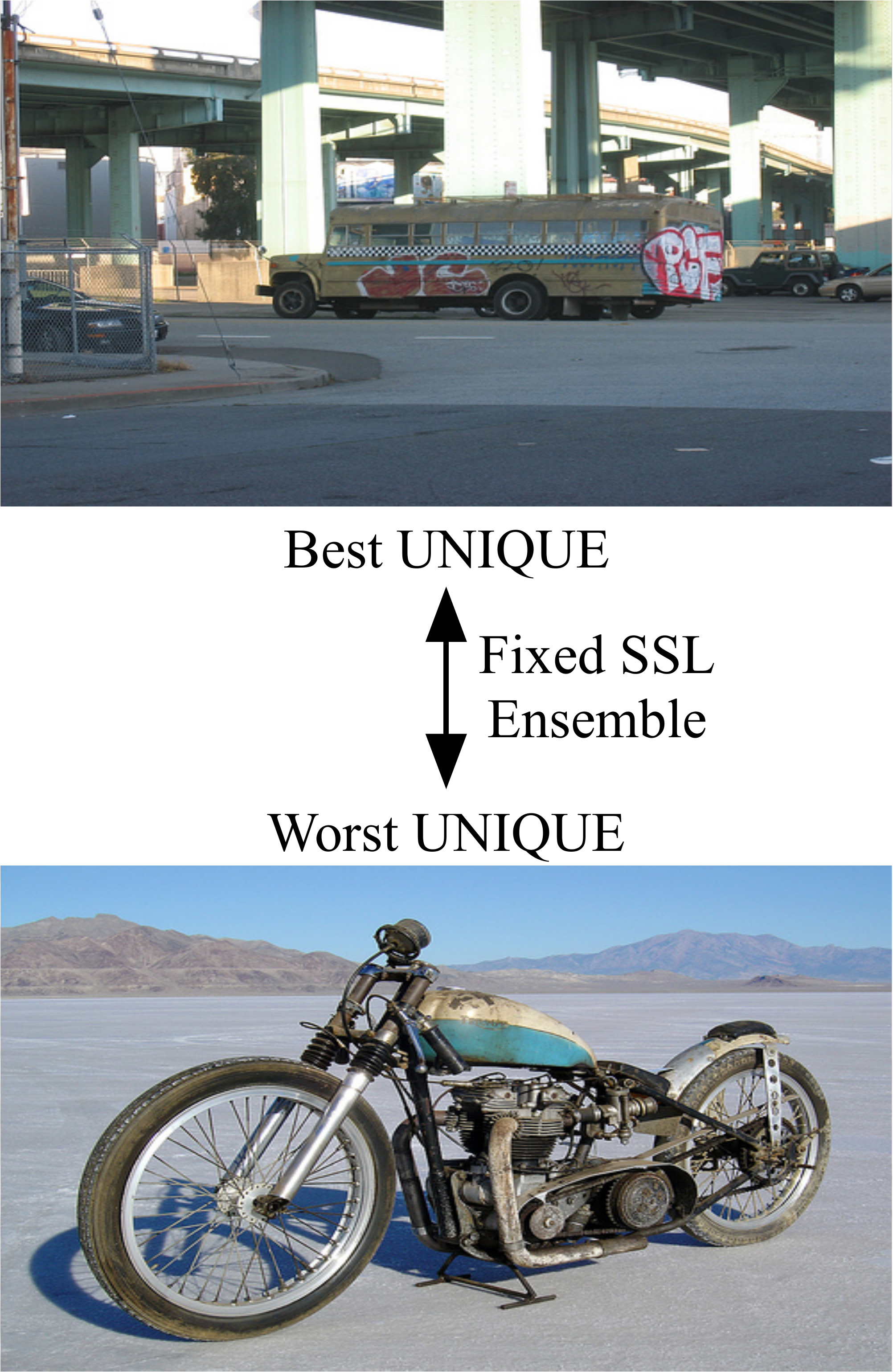}}
	\caption{Representative gMAD pairs between UNIQUE \protect\cite{zhang2020uncertainty} and the SSL ensemble on FLIVE \protect\cite{ying2020patches}. Shown in the sub-caption is the MOS of each image. \textbf{(a)} Fixing UNIQUE at the low-quality level.
    \textbf{(b)}  Fixing UNIQUE at the high-quality level.
    \textbf{(c)} Fixing the SSL ensemble at the low-quality level.
    \textbf{(d)} Fixing the SSL ensemble at the high-quality level.
    }
    \label{fig:gmad_results}
\end{figure*}

\begin{table}[t]
    \caption{Correlation between model predictions and MOSs on the test sets of KonIQ-10k \protect\cite{hosu2020koniq}, SPAQ \protect\cite{fang2020cvpr}, and FLIVE \protect\cite{ying2020patches}, respectively.}
    \label{tab:comparision}
    \vspace{-.3cm}
	\begin{center}
		\begin{tabular}{l|ccc}
    		\toprule[1pt]
			SRCC  & KonIQ-10k & SPAQ & FLIVE  \\
			\hline
			UNIQUE  & $0.860$ & $0.899$ & $0.426$ \\
			Na\"{i}ve Ensemble & $0.855$ & $0.900$ & $0.427$\\
			Joint Ensemble  & $0.864$ & $0.899$ & $0.430$ \\
			SSL Ensemble & $0.861$ & $0.900$ & $0.438$\\
			\hline
			\hline
			PLCC  & KonIQ-10k & SPAQ & FLIVE  \\
		    \hline
			UNIQUE  & $0.859$ & $0.904$ & $0.505$ \\
			Na\"{i}ve Ensemble & $0.859$ & $0.903$ & $0.517$ \\
			Joint Ensemble  & $0.870$ & $0.903$ & $0.521$ \\
			SSL Ensemble  & $0.861$ & $0.904$ & $0.527$  \\
			\bottomrule[1pt]
		\end{tabular}
	\end{center}
\end{table}

\begin{table}[t]
    \caption{Comparison of the failure-spotting efficiency as a function of the number of selected samples on FLIVE \protect\cite{ying2020patches}. A lower correlation coefficient indicates better performance.}
    \label{tab:failurespotting}
    \vspace{-.3cm}
	\begin{center}
		\begin{tabular}{l|cccc}
    		\toprule[1pt]
			SRCC & $500$ & $1,000$ & $1,500$ & $2,000$  \\
			\hline
			UNIQUE  & $0.358$ & $0.373$ & $0.352$ & $0.377$\\
			Na\"{i}ve Ensemble & $0.458$ & $0.397$ & $0.391$ & $0.384$\\
			Joint Ensemble  & $0.346$ & $0.355$ & $0.360$ & $0.370$\\
			SSL Ensemble  & $0.338$ & $0.305$ & $0.318$ & $0.326$ \\
			\hline
			\hline
			PLCC & $500$ & $1,000$ & $1,500$ & $2,000$  \\
			\hline
			UNIQUE  & $0.463$ & $0.460$ & $0.456$ & $0.483$\\
			Na\"{i}ve Ensemble & $0.504$ & $0.461$ & $0.452$ & $0.441$ \\
			Joint Ensemble  & $0.483$ & $0.483$ & $0.488$ & $0.482$ \\
			SSL Ensemble  & $0.378$ & $0.384$ & $0.390$ & $0.400$ \\
			\bottomrule[1pt]
		\end{tabular}
	\end{center}
\end{table}

\vspace{0.3em}\noindent\textbf{Datasets}. We use $60\%$ images in 
 KonIQ-10k \cite{hosu2020koniq} and SPAQ \cite{fang2020cvpr} as the labeled training set, $20\%$ as the validation set, and $20\%$ as the test set, respectively. We treat the entire FLIVE \cite{ying2020patches} as the unlabeled training set. To reduce the bias caused by the random splitting, we repeat the procedure three times, and report the mean results.

\vspace{0.3em}\noindent\textbf{Evaluation Metrics}. We adopt two quantitative criteria: Spearman rank-order correlation coefficient (SRCC) and Pearson linear correlation coefficient (PLCC), to measure prediction performance, respectively. Before computing PLCC, a four-parameter logistic function is suggested in~\cite{antkowiak2000final} to fit model predictions to subjective scores:
\begin{align}
    \hat{f}(x) = (\eta_1 -\eta_2)/(1+\exp(-(f(x)-\eta_3)/|\eta_4|)) + \eta_2,
    \label{eq:nm}
\end{align} 
where $\{\eta_i; i=1,2,3,4\}$ are the parameters to be optimized. 

\subsection{Main Results}
\label{subsec:main_results}
\noindent\textbf{Correlation Results}. We compare our method (termed as the SSL ensemble) against three variants: 1) UNIQUE~\cite{zhang2020uncertainty}, a state-of-the-art BIQA model, corresponding to $M=1$; 2) Na\"{i}ve Ensemble, corresponding to $\lambda=0$ in Eq.~\eqref{eq:jointloss} and $\gamma = 0$ in Eq.~\eqref{eq:tloss}; 3) Joint Ensemble, corresponding to $\gamma = 0$. The mean SRCC and PLCC results are listed in Table~\ref{tab:comparision}, where  we find that the SSL ensemble performs favorably against the competing methods.

\begin{table}[t]
    \caption{The effect of the number of base learners in the SSL ensemble. The default setting is highlighted in bold.}
    \label{tab:number_of_base_learner}
    \vspace{-.3cm}
	\begin{center}
		\begin{tabular}{l|ccc}
    		\toprule[1pt]
			SRCC  & KonIQ-10k & SPAQ & FLIVE  \\
			\hline
			$2$ & $0.859$ & $0.897$ & $0.418$\\
			$4$ & $0.861$ & $0.896$ & $0.421$ \\
			$6$ & $0.861$ & $0.897$ & $0.424$\\
			$\textbf{8}$ & $0.861$ & $0.900$ & $0.438$\\
			$12$ & $0.864$ & $0.900$ & $0.426$\\
			\hline
			\hline
			PLCC  & KonIQ-10k & SPAQ & FLIVE  \\
		    \hline
			$2$ & $0.854$ & $0.901$ & $0.506$\\
			$4$ & $0.862$ & $0.900$ & $0.509$ \\
			$6$ & $0.863$ & $0.900$ & $0.513$\\
			$\textbf{8}$ & $0.861$ & $0.904$ & $0.527$  \\
			$12$ & $0.871$ & $0.903$ & $0.518$\\
			\bottomrule[1pt]
		\end{tabular}
	\end{center}
\end{table}

\vspace{0.3em}\noindent\textbf{gMAD Results}. We next let the SSL ensemble play the gMAD competition game \cite{ma2018group} with UNIQUE on the FLIVE database. Figure~\ref{fig:gmad_results} shows the representative gMAD pairs. It is clear that the pairs of images in (a) and (b), where UNIQUE and the SSL ensemble perform the defender and attacker roles, respectively, exhibit substantially different quality, which is in disagreement with UNIQUE. In contrast, the SSL ensemble correctly predicts the top images to have much better quality than the corresponding bottom images. When switching their roles (see (c) and (d)), we find that the SSL ensemble successfully survives the attacks from UNIQUE, with pairs of images of similar quality according to human perception. This provides a strong indication  that the SSL ensemble leads to better generalization to novel images.

\vspace{0.3em}\noindent\textbf{Failure Identification Results}. It is of great importance to efficiently spot the catastrophic failures of ``top-performing'' BIQA models, offering the opportunity to further improve the models. Wang and Ma \cite{wang2020active} made one of the first attempts to identify the counterexamples of UNIQUE \cite{zhang2020uncertainty} with the help of multiple full-reference IQA models. A significant advantage of the proposed ensemble is that it comes with a natural failure identification mechanism: we may seek samples by the principle of maximal disagreement, as measured by Eq.~\eqref{eq:mse}. This is known as the query by committee \cite{seung1992query} in the active learning literature. To enable comparison with UNIQUE \cite{zhang2020uncertainty}, we re-train it with the dropout technique \cite{srivastava2014dropout} and use Monte Carlo dropout~\cite{gal2016dropout} to approximate Eq.~\eqref{eq:mse} for failure identification. 
Table~\ref{tab:failurespotting} lists the SRCC and PLCC results as a function of the sample size on FLIVE. Here lower correlation indicates higher failure-spotting efficiency. As can be seen, the SSL ensemble achieves the lowest correlation across all sample sizes, providing strong justifications of exploiting unlabeled data for diversity promotion. We are currently testing the use of the spotted samples for model refinement in a similar active fine-tuning framework described in \cite{wang2020active}. Preliminary results show that the samples with the maximal disagreement are beneficial for improving the model generalizability.

\begin{table}[t]
    \caption{The effect of the trade-off parameter ($\gamma$ in Eq.~\eqref{eq:tloss}) for training the SSL ensemble.}
    \label{tab:tradeoff_gamma}
    \vspace{-.3cm}
	\begin{center}
		\begin{tabular}{l|ccc}
    		\toprule[1pt]
			SRCC  & KonIQ-10k & SPAQ & FLIVE  \\
			\hline
			$0.00$ & $0.855$ & $0.900$ & $0.427$\\
			$0.04$ & $0.864$ & $0.899$ & $0.429$\\
			$\textbf{0.06}$ & $0.861$ & $0.900$ & $0.438$\\
			$0.08$ & $0.860$ & $0.895$ & $0.432$\\
			$0.10$ & $0.851$ & $0.894$ & $0.437$\\
			\hline
			\hline
			PLCC  & KonIQ-10k & SPAQ & FLIVE  \\
		    \hline
		    $0.00$ & $0.870$ & $0.903$ & $0.521$\\
			$0.04$ & $0.868$ & $0.903$ & $0.518$ \\
			$\textbf{0.06}$ & $0.861$ & $0.904$ & $0.527$\\
			$0.08$ & $0.865$ & $0.899$ & $0.522$\\
			$0.10$ & $0.848$ & $0.897$ & $0.525$\\
			\bottomrule[1pt]
		\end{tabular}
	\end{center}
\end{table}

\begin{table}[t]
    \caption{The effect of different splitting points in the SSL ensemble using ResNet-18 as the backbone. The configuration in the first column indicates the layer, after which the computation is not shared.}
    \label{tab:splitting_point}
    \vspace{-.3cm}
	\begin{center}
		\begin{tabular}{l|ccc}
    		\toprule[1pt]
			SRCC  & KonIQ-10k & SPAQ & FLIVE  \\
			\hline
			First Convolution  & $0.866$ & $0.899$ & $0.434$\\
			Residual Block $1$ & $0.865$ & $0.899$ & $0.434$\\
			Residual Block $2$ & $0.864$ & $0.899$ & $0.432$\\
			\textbf{Residual Block} $\textbf{3}$ & $0.861$ & $0.900$ & $0.438$\\
			Residual Block $4$ & $0.864$ & $0.900$ & $0.437$\\
			\hline
			\hline
			PLCC  & KonIQ-10k & SPAQ & FLIVE  \\
		    \hline
			First Convolution   & $0.871$ & $0.903$ & $0.525$\\
			Residual Block $1$  & $0.869$ & $0.903$ & $0.525$\\
			Residual Block $2$  & $0.869$ & $0.903$ & $0.523$\\
			\textbf{Residual Block} $\textbf{3}$  & $0.861$ & $0.904$ & $0.527$\\
			Residual Block $4$  & $0.864$ & $0.904$ & $0.525$\\
			\bottomrule[1pt]
		\end{tabular}
	\end{center}
\end{table}

\subsection{Ablation Studies} 
In this subsection, we perform ablation experiments to examine the influence of three hyper-parameters: 1) the number of base learners, $M$, 2) the trade-off parameter for ensemble accuracy and diversity, $\bm{\gamma}$, and 3) the splitting point, before which all base learners share the computation.

Table~\ref{tab:number_of_base_learner} shows the effect of employing different numbers of base learners (\textit{i.e.}, $M\in\{2,4,6,8,12\}$) in the SSL ensemble, where we find that our method is fairly stable with regard to this hyperparameter, and adding more base learners slightly increase the correlation numbers.  We next test the influence of the trade-off parameter $\gamma$ in Eq.~\eqref{eq:tloss}. 
Table~\ref{tab:tradeoff_gamma} shows the performance change with $\gamma$ sampled from $\{0.00, 0.04, 0.06, 0.08, 0.10\}$. It is not surprising that a larger value (\textit{e.g.}, $\gamma=0.10$) may harm the ensemble performance, especially on KonIQ-10k.
This is because excessive diversity may destroy the prediction accuracy of base learners and (subsequently) the ensemble \cite{brown2010good}. This phenomenon would be more pronounced if the adopted diversity measure is unbounded from above (\textit{e.g.}, Eq.~\eqref{eq:mse}). In general, the correlation is not sensitive when $\gamma$ is relatively small. Last, we explore the influence of the splitting point determining how much computation is shared by the base learners. From the Table~\ref{tab:splitting_point},  we observe stable performance at different splitting points. Therefore, it is encouraged to share more layers to reduce the inference time and the memory requirement.

\section{Conclusions}
\label{sec:conclusion}
In this paper, we have conducted  an empirical study to probe the advantages of incorporating unlabeled data for training BIQA ensembles. This naturally leads to a semi-supervised formulation, where we maximized the ensemble accuracy on labeled data and the ensemble diversity on unlabeled data. Through comprehensive experiments, we arrived at three interesting findings. First, the diversity-driven SSL ensemble does not achieve correlation improvements on existing IQA databases. Second, despite similar correlation performance, our ensemble shows much stronger generalizability in the gMAD competition with greater potentials for use in monitoring real-world quality problems. Third, our ensemble has a built-in failure-identification mechanism with demonstrated efficiency. This points to an interesting avenue for future work - active fine-tuning BIQA models in the semi-supervised setting.
\bibliographystyle{named}
\bibliography{ijcai21}

\end{document}